\documentclass[runningheads]{llncs}
\usepackage[T1]{fontenc}
\usepackage{graphicx}
\usepackage{multirow}
\usepackage{amsmath}
\usepackage{color}
\begin{document}
\title{Vision-Based Driver Drowsiness Monitoring: Comparative Analysis of YOLOv5–v11 Models}
%
%
\author{Dilshara Herath\inst{1}\orcidID{0009-0004-1071-1698} \and
Chinthaka Abeyrathne\inst{1}\orcidID{0009-0005-7911-723X} \and
Prabhani Jayaweera\inst{2}\orcidID{
0009-0003-9266-716X}
}
\authorrunning{D. Herath et al.}
%
\institute{Electrical and Information Engineering\inst{1}, Computer Engineering\inst{2}, University of Ruhuna, Sri Lanka 
\email{\{dilshara.herath3, chinthaka.edu.ruh, prabhanijayaweera\}@gmail.com}\\
}
\maketitle              
\begin{abstract}
Driver drowsiness remains a critical factor in road accidents, accounting for thousands of fatalities and injuries each year. This paper presents a comprehensive evaluation of real-time, non-intrusive drowsiness detection methods, focusing on computer vision based YOLO (You Look Only Once) algorithms. A publicly available dataset namely, UTA-RLDD was used, containing both awake and drowsy conditions, ensuring variability in gender, eyewear, illumination, and skin tone. Seven YOLO variants (v5s, v9c, v9t, v10n, v10l, v11n, v11l) are fine-tuned, with performance measured in terms of Precision, Recall, mAP@0.5, and mAP@ 0.5–0.95. Among these, YOLOv9c achieved the highest accuracy (mAP@ 0.5 = 0.986, Recall = 0.978) while YOLOv11n strikes the optimal balance between precision (0.954) and inference efficiency, making it highly suitable for embedded deployment. Additionally, we implement an Eye Aspect Ratio (EAR) approach using Dlib’s facial landmarks, which despite its low computational footprint exhibits reduced robustness under pose variation and occlusions. Our findings illustrate clear trade-offs between accuracy, latency, and resource requirements, and offer practical guidelines for selecting or combining detection methods in autonomous driving and industrial safety applications.

\end{abstract}
\section{Introduction}

Road safety agencies and researchers widely recognize driver drowsiness as a serious hazard in transportation. Drowsy or fatigued driving has been implicated in a substantial number of traffic accidents, injuries, and fatalities each year. For example, the U.S. National Highway Traffic Safety Administration (NHTSA) reported that in 2017 alone approximately 91,000 police-reported crashes were caused by drowsy drivers, leading to about 50,000 injuries and nearly 800 deaths\cite{s22052069}. The societal cost of drowsy driving is enormous one analysis put the economic loss at roughly \$12.5 billion in a single year. Equally important are the human costs: surveys indicate drowsiness while driving is widespread, as 58\% of drivers in one study admitted to driving while fatigued and 14.5\% even reported dozing off at the wheel \cite{nature1}. Studies have found that driver drowsiness levels can be even higher during automated driving than manual driving\cite{nature2}. Monitoring driver alertness remains critical even advanced intelligent vehicles require a vigilant driver for take-over requests in emergency situations. Therefore, developing accurate and real-time drowsiness detection technology is essential for enhancing road safety and saving lives.

Over the past decade, a variety of approaches have been explored to detect driver drowsiness before an accident occurs. One class of methods monitors the driver’s biological signals such as brain activity recognition using electroencephalogram (EEG), eye movements electrooculography  (EOG), heart rate, or other biosignals to infer drowsiness\cite{8718312}\cite{8520803}\cite{10603125}. These physiological measures are proven to be highly sensitive to early drowsiness-related changes, often capturing subtle signs before any outward behavior appears. For instance, EEG-based systems can detect shifts in brain wave patterns that correlate with fatigue, enabling early warnings before the driver visibly nods off. Experimental EEG systems have achieved high detection accuracies (often 90\%+) by recognizing these initial neural or ocular cues of fatigue\cite{s22052069}. However, physiological methods require specialized sensors attached to the driver, which poses practical limitations. Many EEG/EOG-based prototypes involve electrodes on the scalp or face, which are intrusive for drivers to wear during regular driving.

Another category relies on vehicular performance data and driving patterns to infer drowsiness. These methods monitor signals such as steering wheel movements (steering wheel angle – SWA), lane positioning (lane departures or weaving), speed variance, and other driving behaviors under the assumption that a drowsy driver will exhibit erratic control of the vehicle \cite{9509912} \cite{9929437}. However, these techniques by themselves are generally the least reliable for detecting drowsiness. The primary reason is that changes in driving performance tend to appear later in the progression of drowsiness, often when the driver is already quite impaired. Critical early signs like prolonged eye closures or micro-sleeps may not immediately translate into detectable vehicle deviations. 

One of the most active research areas is computer vision-based drowsiness detection, which monitors the driver’s facial features, eyes, and head movements via a camera \cite{9435480} \cite{safarov2023real} \cite{9087013}. These image-based methods aim to detect observable behavioral cues of fatigue for example, a driver’s eyes closing or blinking slowly, frequent yawning, head nodding, or gaze fixation loss. Such cues are strongly correlated with drowsiness; indeed, visible signs like repeated eye closures and yawning typically emerge before a vehicle wanders out of its lane. Vision-based systems have several practical advantages: they are contactless, non-intrusive, and increasingly cost-effective. A simple camera (e.g. an IR webcam or dashboard camera) is used, so drivers do not need to wear any equipment. Despite these strengths, vision-based approaches face notable challenges that can affect their reliability in practice. Their performance can degrade under variable lighting (e.g. nighttime driving or glare from sunlight). However, the application of YOLO and similar state-of-the-art vision models to drowsiness detection is still an emerging area, and open questions remain regarding how different architectures compare and how well they generalize. The main contributions of this paper are:

\begin{itemize}
    \item Diverse Dataset Curation: We curate and annotate a novel image dataset from the public UTA-RLDD video corpus, ensuring balanced representation across “awake” and “drowsy” states and diversity in gender, eyewear use, illumination conditions, and skin tones. 
    \item Comprehensive YOLO Variant Evaluation: We fine-tune and benchmark seven YOLO architectures (v5s, v9c, v9t, v10n, v10l, v11n, v11l) under a uniform training protocol, providing the first systematic comparison of their precision, recall, mAP@ 0.5, and mAP@ 0.5–0.95 metrics on a driver drowsiness task.

\end{itemize}

The rest of the paper is organized as follows: Section III explains the system architecture; Section IV, experimental results with discussion and finally Section V, conclusions are drawn with the future research directions.

\section{System Architecture}

\begin{figure}[htbp]  
    \centering
    \includegraphics[width=0.95\textwidth]{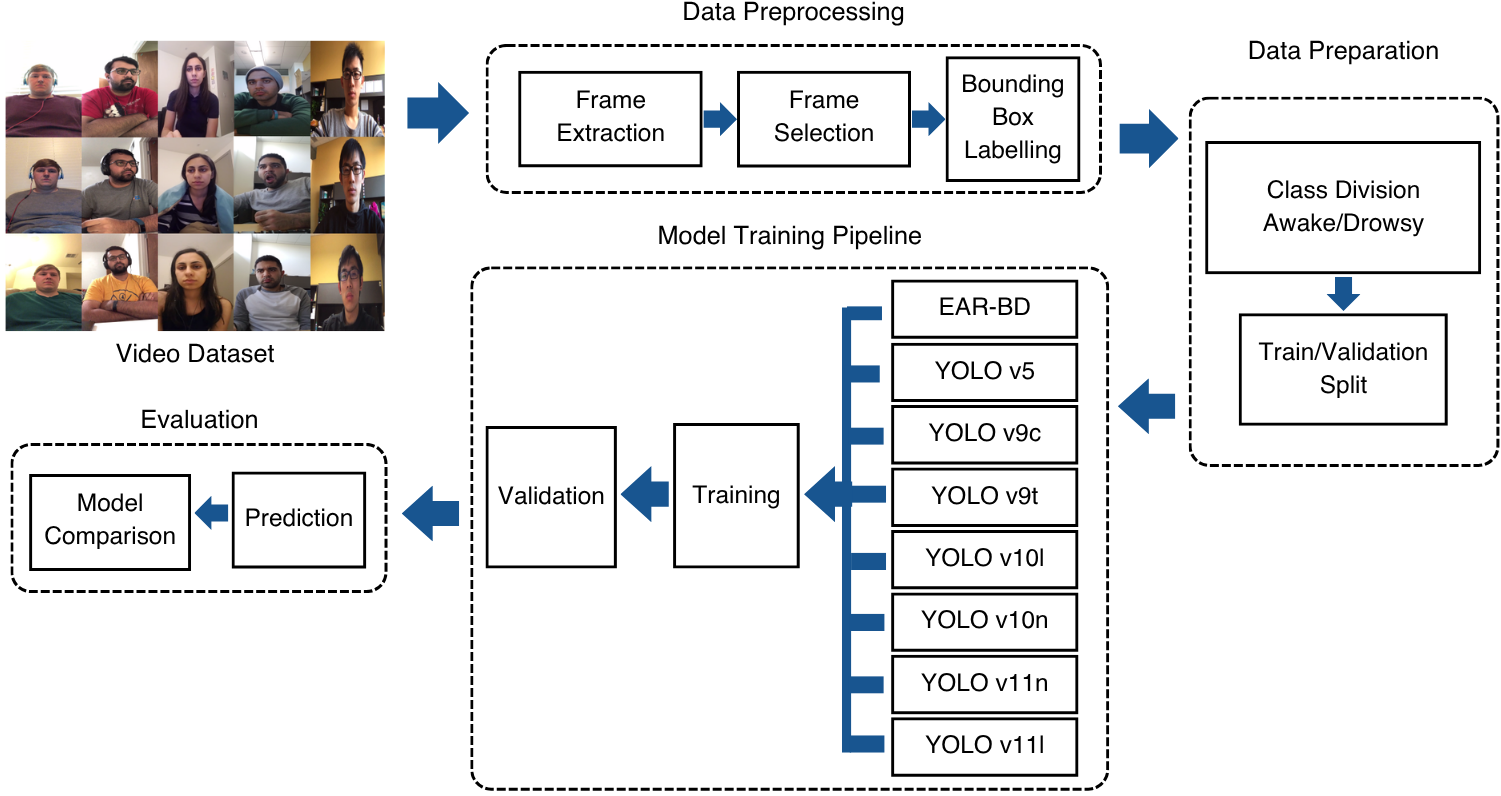}
    \caption{The complete system architecture of the Drowsiness Detection System}
    \label{sysarchi}
\end{figure}

This section explains the details of the system architecture of the drowsiness detection system we have implemented. The Fig. \ref{sysarchi} shows the complete system. First the dataset is preprocessed by a pipeline, which includes Frame extraction, frame selection followed by annotation using bounding box labeling. Then data splitting was performed before the images were fed to the computer vision pipeline. The dataset then trained and evaluated using different YOLO models namely, v5s, v9c, v9t, v10n, v10l, v11n, v11l and a classical approach. Finally, the prediction is performed for model comparison. 

The dataset used for the system is the Arlington Real-Life Drowsiness Dataset (UTA-RLDD). As described by Ghoddoosian et al. \cite{ghoddoosian2019realisticdatasetbaselinetemporal}, it comprises approximately 30 hours of high-resolution RGB video footage recorded, at a frame rate of 30 frames per second (FPS), from 60 healthy volunteers under three vigilance states: fully alert, low vigilance, and drowsy. From this larger cohort, we purposefully selected 18 participants (totaling 54 video sequences) to ensure diversity across gender, eyewear usage, ambient illumination (low- and high-light conditions), and a range of skin tones. This balanced subset was chosen to challenge the model with real-world variability and to promote robust generalization across different driver profiles and environmental settings.

To transform the video data into a suitable format for analysis, frames were extracted at a rate of 30 FPS (Frames per second) and saved as jpg images. This process yielded 3418 total frames, ensuring a balanced representation of drowsy and awake conditions. Image annotation was carried out using the \textit{MakeSense.ai} platform to generate bounding-box labels for subsequent object-detection training. We manually drew rectangular boxes around each driver’s face, ensuring full coverage of critical regions such as the eyes and mouth. This precise localization enables the detection network to focus on the most salient visual features associated with drowsiness. Each face bounding box was assigned one of two mutually exclusive class labels: “Drowsy” (indicating observable signs of fatigue, such as prolonged eye closure or yawning) or “Awake” (indicating normal, alert behavior). Upon completion and verification, annotation files were exported in the YOLO text format. For each image, a corresponding .txt file was generated, containing one line per bounding box with normalized coordinates (x\_center, y\_center, width, height) and the integer class index. This resulted, 1476 and 1942 images for awake and drowsy classes respectively. 

The fully annotated image dataset was then partitioned into training (70\%), validation (20\%), and test (10\%) sets using a stratified sampling strategy that preserved the original “drowsy” versus “awake” distribution in each subset. All splits were randomized with a fixed seed (42) to guarantee reproducibility. 

\subsection{YOLO model Implementation}

In this work, we leveraged the YOLO series of one-stage detectors due to their unparalleled balance of speed and accuracy in real-time object detection. Table \ref{tab:yolo_comp} illustrates the YOLO variants we chose to span different complexity–performance regimes. 

\begin{table}[]
\centering
\caption{Different YOLO models used for drowsiness detection}
\label{tab:yolo_comp}
\begin{tabular}{|l|c|c|l|}
\hline
\multicolumn{1}{|c|}{\textbf{Model}} & \textbf{params(M)} & \textbf{FLOPs (B)} & \multicolumn{1}{c|}{\textbf{Description}} \\ \hline
YOLOv5s  & 9.1  & 24.0    & Small version, fast inference \\ \hline
YOLOv9c  & 25.5 & 102.8 & Compact version of YOLOv9     \\ \hline
YOLOv9t  & 2.0    & 7.7   & Tiny version of YOLOv9        \\ \hline
YOLOv10n & 2.3  & 6.7   & Nano version of YOLOv10       \\ \hline
YOLOv10l & 24.4 & 120.3 & Large version of YOLOv10      \\ \hline
YOLOv11n & 2.6  & 6.5   & Nano version of YOLOv11       \\ \hline
YOLOv11l & 25.3 & 86.9  & Tiny version of YOLOv11       \\ \hline
\end{tabular}
\end{table}

\subsubsection{Bounding‐Box Parameterization}

In YOLO’s grid‐based detection paradigm, the input image is divided into an \(S\times S\) grid, and each grid cell \((i,j)\) is responsible for predicting \(B\) candidate bounding boxes. For each box, the network outputs four offset parameters \((t_x,t_y,t_w,t_h)\), along with a confidence score and class probabilities. These offsets are decoded into absolute box center coordinates \((b_x,b_y)\) and dimensions \((b_w,b_h)\) as shown in Equation~\eqref{eq:decode},
\begin{equation}
\begin{aligned}
b_x &= \sigma(t_x) + c_x,\\
b_y &= \sigma(t_y) + c_y,\\
b_w &= p_w\,e^{t_w},\\
b_h &= p_h\,e^{t_h},
\end{aligned}
\label{eq:decode}
\end{equation}

where, \(\sigma(\cdot)\) denotes the sigmoid activation, constraining the predicted center offsets to the interval \((0,1)\) relative to the grid cell; \((c_x,c_y)\) represents the top‐left corner of cell \((i,j)\) in grid coordinates; and \((p_w,p_h)\) are the predefined anchor‐box dimensions for that cell which scales the anchor box by the learned exponential factors to model variable object sizes. Finally, to convert these values back into the original image coordinate system, the centers \((b_x,b_y)\) are divided by \(S\) and the dimensions \((b_w,b_h)\) by the full image width and height, respectively. This parameterization supports stable regression during training and precise localization at inference.

\subsubsection{Confidence Score}
Each predicted box has an associated confidence score, given asin Eq. ~\eqref{eq:confidence}

\begin{equation}
\hat{C} = \sigma(t_c) \approx \Pr(\mathrm{Object}) \times \mathrm{IoU}_{\mathrm{pred},\mathrm{truth}},
\label{eq:confidence}
\end{equation}

where, $t_c$ is the raw network output and $\mathrm{IoU}$ is the Intersection over Union between the predicted and ground-truth boxes.

\subsubsection{Classification Probabilities}
For each grid cell containing an object, class probabilities are predicted as in Eq. ~\eqref{eq:classprob}
\begin{equation}
\hat{p}_k = \Pr(\mathrm{class}=k \mid \mathrm{Object}),\quad k=1,\dots,K.
\label{eq:classprob}
\end{equation}

\subsubsection{YOLO Loss Function}
YOLO’s multi-component loss combines localization, confidence, and classification terms shows in Eq. ~\eqref{eq:yololoss}
\begin{equation}
\begin{aligned}
\mathcal{L} =\; &\lambda_{\mathrm{coord}}
\sum_{i=1}^{S^2}\sum_{j=1}^{B} \mathbf{1}_{ij}^{\mathrm{obj}}
\bigl[(b_x - \hat b_x)^2 + (b_y - \hat b_y)^2\bigr] \\
+\, &\lambda_{\mathrm{coord}}
\sum_{i=1}^{S^2}\sum_{j=1}^{B} \mathbf{1}_{ij}^{\mathrm{obj}}
\bigl[(\sqrt{b_w} - \sqrt{\hat b_w})^2 + (\sqrt{b_h} - \sqrt{\hat b_h})^2\bigr] \\
+\, &\sum_{i=1}^{S^2}\sum_{j=1}^{B}\mathbf{1}_{ij}^{\mathrm{obj}}
\bigl(C_{ij} - \hat C_{ij}\bigr)^2 \\
+\, &\lambda_{\mathrm{noobj}}
\sum_{i=1}^{S^2}\sum_{j=1}^{B}\mathbf{1}_{ij}^{\neg\mathrm{obj}}
\bigl(C_{ij} - \hat C_{ij}\bigr)^2 \\
+\, &\sum_{i=1}^{S^2} \mathbf{1}_{i}^{\mathrm{obj}}
\sum_{k=1}^{K} \bigl(p_{i}(k) - \hat p_{i}(k)\bigr)^2,
\end{aligned}
\label{eq:yololoss}
\end{equation}
where $\mathbf{1}_{ij}^{\mathrm{obj}}$ is 1 if object appears in cell $i$ for box $j$, and $\mathbf{1}_{ij}^{\neg\mathrm{obj}}$ is 1 otherwise. Hyperparameters $\lambda_{\mathrm{coord}}$ and $\lambda_{\mathrm{noobj}}$ balance localization and confidence penalties.

\subsubsection{YOLO Model Training}

All YOLO variants were fine-tuned under an identical training regimen to ensure a fair and reproducible comparison. Training was carried out for 50 epochs on Google Colab using an NVIDIA Tesla T4 (L4) GPU with 16 GB of VRAM. Input images were uniformly resized to 640×640 pixels and processed in batches of 16. We initialized each network with pretrained MS COCO weights and optimized using stochastic gradient descent with a momentum of 0.937 and weight decay of 0.0005. The learning rate followed a cosine-annealing schedule, decaying from 0.01 to 0.0001 over the course of training. To improve robustness and generalization, on-the-fly data augmentations including Mosaic composition, MixUp, random horizontal flips, and brightness/contrast jitter were applied. Throughout training, we monitored box-regression, objectness, and classification losses on the training set, and evaluated precision, recall, mAP@ 0.5, and mAP@ 0.5:0.95 on the validation set after each epoch. Upon completion, the best model weights those achieving the highest validation mAP were preserved for subsequent testing and analysis.

\subsubsection{Evaluation Metrics}

To rigorously assess detector performance, we employ the following standard object‐detection metrics:

\begin{itemize}
  \item \textbf{Precision:}  
    The ratio of true positive detections to all positive predictions, measuring the false‐alarm rate:
    \begin{equation}
      \mathrm{Precision} = \frac{\mathrm{TP}}{\mathrm{TP} + \mathrm{FP}}
      \label{eq:precision}
    \end{equation}

  \item \textbf{Recall:}  
    The ratio of true positive detections to all actual objects, indicating the model’s ability to find all instances:
    \begin{equation}
      \mathrm{Recall} = \frac{\mathrm{TP}}{\mathrm{TP} + \mathrm{FN}}
      \label{eq:recall}
    \end{equation}

  \item \textbf{Average Precision (AP):}  
    The area under the Precision–Recall curve for a given Intersection‐over‐Union (IoU) threshold, defined as:
    \begin{equation}
      \mathrm{AP} = \int_{0}^{1} p(r)\,\mathrm{d}r
      \label{eq:ap}
    \end{equation}
    where \(p(r)\) is precision as a function of recall \(r\).

  \item \textbf{mAP@0.5:}  
    The mean of AP values computed at a fixed IoU threshold of 0.50 across all \(C\) classes:
    \begin{equation}
      \mathrm{mAP}_{@0.5} = \frac{1}{C} \sum_{c=1}^{C} \mathrm{AP}_{c}^{\mathrm{IoU}=0.5}
      \label{eq:map50}
    \end{equation}

  \item \textbf{mAP@0.5–0.95:}  
    The mean of AP values averaged over multiple IoU thresholds \(t \in \{0.50,0.55,\ldots,0.95\}\) and all \(C\) classes:
    \begin{equation}
      \mathrm{mAP}_{@0.5\text{--}0.95}
      = \frac{1}{C\,|\mathcal{T}|}
        \sum_{c=1}^{C}
        \sum_{t \in \mathcal{T}}
        \mathrm{AP}_{c}^{\mathrm{IoU}=t},
      \quad
      \mathcal{T} = \{0.50,0.55,\ldots,0.95\}
      \label{eq:map5095}
    \end{equation}
\end{itemize}

\subsection{EAR-BD Method for Drowsiness Detection}

To compare the Yolo model's performances, we implemented a lightweight, real-time mechanism namely, EAR-BD (Eye Aspect Ratio Based Drowsiness Detection) for inferring driver drowsiness by monitoring eyelid movement using only a standard camera and CPU-based facial landmark detection.

\begin{figure}[htbp]  
    \centering
    \includegraphics[width=0.55\textwidth]{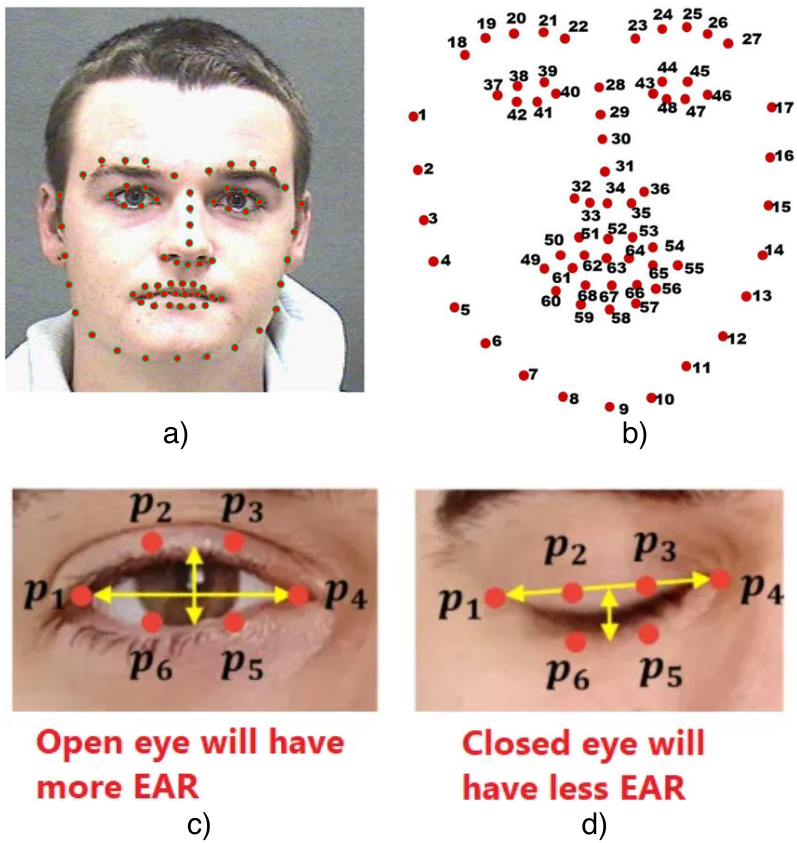}
    \caption{Identification of facial landmarks using Dlib. a) Facial landmarks. b)
The position and order of 68 points on the face and EAR values comparison for c) open eyes d) closed eyes}
    \label{earbd}
\end{figure}

This approach leverages Dlib’s pretrained 68-point shape predictor to locate eye landmarks in each  frame as shown in Fig. \ref{earbd} a) and b). Then, it computes a simple geometric ratio of vertical to horizontal eye opening, and applies a threshold over consecutive frames to flag potential microsleep events (Refer to Fig. \ref{earbd} c) and d)).

For each eye, the EAR is calculated by taking the ratio of the sum of two vertical distances to the horizontal distance between eye corners as in Eq. ~\eqref{eq:ear},
\begin{equation}
\mathrm{EAR} = \frac{\lVert p_2 - p_6 \rVert + \lVert p_3 - p_5 \rVert}{2\lVert p_1 - p_4 \rVert}
\label{eq:ear}
\end{equation}

where, \(\lVert.\rVert\) denotes the Euclidean distance between landmark coordinates. As the eyelid closes, the numerator decreases, causing EAR to drop towards zero. Typical thresholds for eye closure detection range from 0.15 to 0.25, empirically determined for each camera setup and driver population.

A frame is classified as ``eye-closed'' if its computed EAR falls below a threshold \(\tau\). To distinguish brief blinks from sustained eye closure indicative of drowsiness, we count the number of consecutive frames \(N_{\text{close}}\) for which \(\mathrm{EAR} < \tau\). When this count exceeds a predefined limit \(N_{\mathrm{thresh}}\), the system issues a drowsiness alert. Formally, the drowsiness flag is defined as in Eq. ~\eqref{eq:drowsy_flag},
\begin{equation}
\text{drowsy\_flag} =
\begin{cases}
1, & N_{\text{close}} \;\ge\; N_{\mathrm{thresh}},\\
0, & \text{otherwise}.
\end{cases}
\label{eq:drowsy_flag}
\end{equation}

\section{Experimental Results}

The below Table \ref{tab:final}. summarizes the detection performance (Precision, Recall, mAP@ 0.5, mAP@ 0.5–0.95) and average epoch training time for seven YOLO variants and EAR-BD method. A detailed examination of these results reveals important trade-offs in accuracy, robustness, and computational cost, key considerations for real-time drowsiness monitoring in autonomous vehicles or other safety-critical systems.

\begin{table}[]
\centering
\caption{Comparison of YOLO models (v5s, v9c, v9t, v10n, v10l, v11n, v11l) and EAR-BD method performance on drowsiness detection }
\label{tab:final}
\begin{tabular}{|l|c|c|c|c|c|}
\hline
\multicolumn{1}{|c|}{\textbf{Model}} &
  \textbf{Precision} &
  \textbf{Recall} &
  \textbf{mAP50} &
  \textbf{mAP50-95} &
  \textbf{Avg. epoch time} \\ \hline
EAR-BD    & 0.812 & 0.82  & N/A   & N/A   & N/A      \\ \hline
YOLOv5s  & 0.920 & 0.904 & 0.962 & 0.761 & N/A      \\ \hline
YOLOv9c  & 0.934 & 0.978 & 0.986 & 0.800 & 1000.013 \\ \hline
YOLOv9t  & 0.954 & 0.941 & 0.974 & 0.792 & 452.890  \\ \hline
YOLOv10n & 0.933 & 0.935 & 0.967 & 0.778 & 315.413  \\ \hline
YOLOv10l & 0.911 & 0.947 & 0.963 & 0.770 & 1083.836 \\ \hline
YOLOv11n & 0.954 & 0.941 & 0.981 & 0.795 & 273.718  \\ \hline
YOLOv11l & 0.909 & 0.975 & 0.977 & 0.789 & 886.343  \\ \hline
\end{tabular}
\end{table}

Among all variants, YOLOv9c attains the highest detection performance, achieving a Precision of 0.934, Recall of 0.978, mAP@0.5 of 0.986, and mAP@0.5– 0.95 of 0.800. This indicates that YOLOv9c not only correctly identifies nearly all true drowsiness events (high Recall) but also maintains few false alarms (high Precision) and robust localization across multiple IoU thresholds. However, its substantial computational cost over 1,000 s per epoch suggests that deploying YOLOv9c for real-time inference would require powerful hardware or cloud offloading. 

The nano and tiny variants have a different balance. YOLOv11n delivers a Precision of 0.954, Recall of 0.941, mAP@ 0.5 of 0.981, and mAP@ 0.5–0.95 of 0.795, while completing each training epoch in only 273.7 s. This combination of near–state-of-the-art accuracy and minimal latency makes YOLOv11n particularly attractive for embedded or edge applications, where both model compactness and inference speed are paramount. Similarly, YOLOv9t achieves the highest Precision (0.954) with a respectable Recall (0.941) and moderate training time (452.9 s), favoring scenarios in which false positives must be strictly controlled.

\begin{figure}[htbp]  
    \centering
    \includegraphics[width=0.9\textwidth]{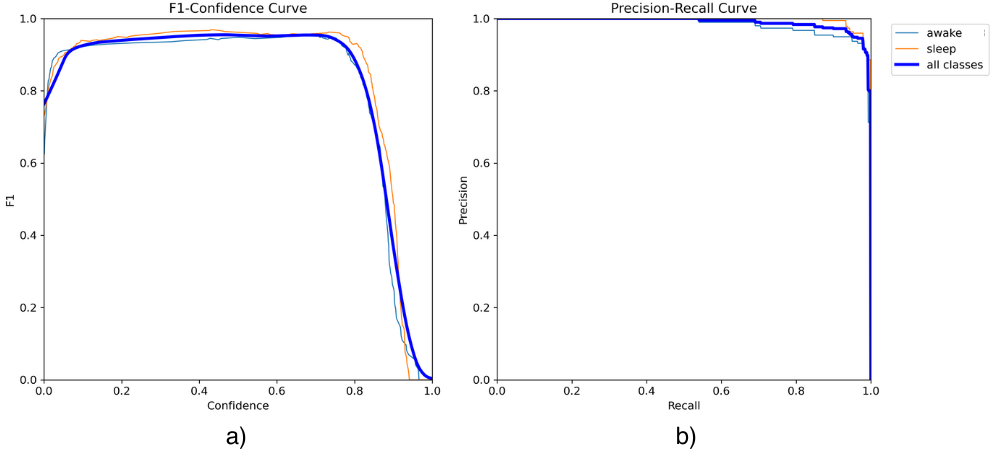}
    \caption{Performance Graphs of a) F1–confidence b) Precision-Recall curve for YOLOv9c model}
    \label{yolo10}
\end{figure}

YOLOv10n offers a balanced trade-off, with Precision and Recall both around 0.933–0.935, mAP@0.5 of 0.967, and a fast average epoch time of 315.4 s. This suggests that for deployments on highly resource-constrained platforms such as inexpensive driver-monitoring modules YOLOv10n can achieve acceptable detection performance while preserving low overhead. The baseline YOLOv5s model, while fast and lightweight, trails newer variants with a Precision of 0.920, Recall of 0.904, and mAP@0.5–0.95 of 0.761. This underscores the advancements realized in the YOLOv9–v11 families, which leverage modern backbone architectures and enhanced training strategies to boost both accuracy and efficiency. 

\begin{figure}[htbp]  
    \centering
    \includegraphics[width=0.7\textwidth]{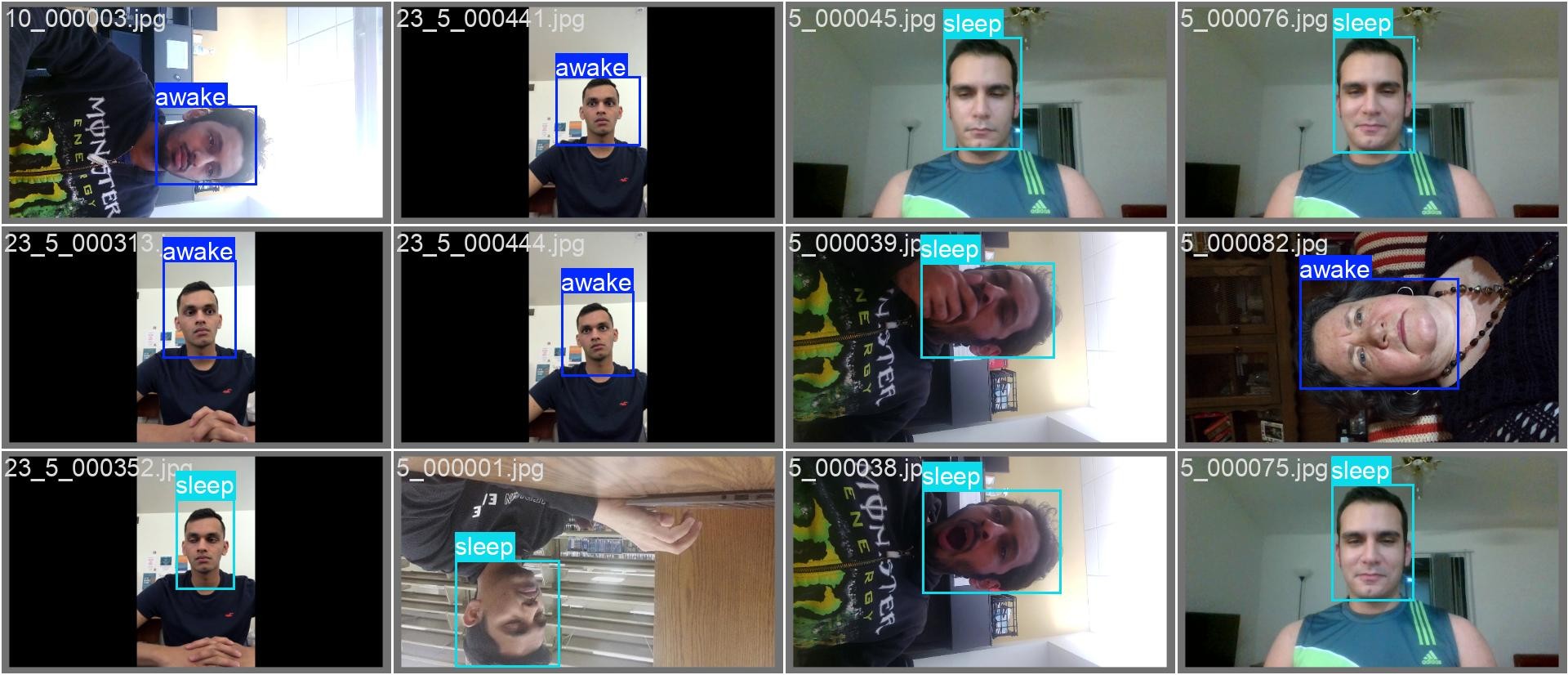}
    \caption{Sample batch of images predicted using the YOLOv9c model}
    \label{batchyolo}
\end{figure}

Fig. \ref{yolo10} shows a performance graphs of F1–confidence and Precision-Recall curve for YOLOv9c model. The F1–confidence curves reveal that both the “awake” (blue) and “sleep” (orange) classes achieve their highest harmonic mean of precision and recall at a confidence threshold of approximately 0.46, where the combined macro-F1 peaks at 0.96 making this threshold ideal for balancing false positives and false negatives in deployment. Meanwhile, the precision–recall plot demonstrates exceptionally high precision (> 0.95) maintained across most recall values, with class-specific average precisions of 0.981 for awake and 0.991 for sleep and an overall mAP@0.50 of 0.986. Together, these curves confirm that the model not only localizes and classifies drowsiness signals with outstanding accuracy but also offers a clear, data driven threshold for decision-making, ensuring both robustness and reliability in real-time autonomous-vehicle monitoring. Fig. \ref{batchyolo} shows 
sample batch of images predicted using the YOLOv9c model. 

Unlike deep-learning detectors, EAR-BD is entirely non-intrusive and runs efficiently on embedded processors, making it well suited for low-cost or resource-constrained applications such as aftermarket driver-monitoring modules. When compared to our YOLO models, which achieved mAP@0.5 scores between 0.962 (YOLOv5s) and 0.986 (YOLOv9c) on the same dataset, the EAR approach exhibits lower overall reliability and is limited to capturing only ocular cues, whereas object‐detection networks can learn to recognize a broader set of drowsiness indicators (e.g., yawning, head nodding) via bounding‐box classification. Also, the EAR method’s reliance on a fixed threshold makes it inherently sensitive to head‐pose variations, partial occlusions (e.g., eyeglasses), and fluctuating illumination, often necessitating manual recalibration for different drivers or lighting conditions. The Fig. \ref{earbd} shows how EAR-BD has detected the eye lids and predicted the state of the person, whether he/she is drowsy or not.

\begin{figure}[htbp]  
    \centering
    \includegraphics[width=0.65\textwidth]{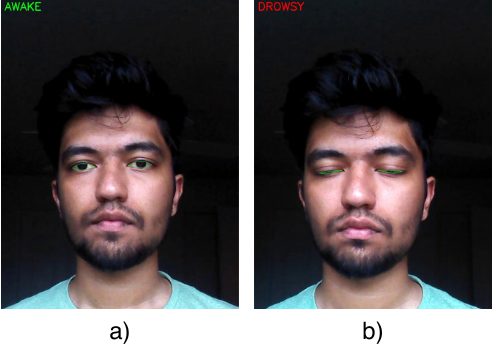}
    \caption{EAR-BD method detecting drowsy and awake persons}
    \label{earbd}
\end{figure}

Our systematic experiments demonstrated that the compact YOLOv11n model achieves a nearly state-of-the-art detection performance with minimal latency, making it especially well suited for embedded, on-board deployment. Furthermore, YOLOv9c attained the strongest overall accuracy yet with higher computational demands. 

\section{Conclusion}

In this study, we have investigated the efficacy of modern computer‐vision techniques for real‐time driver drowsiness detection, with a particular focus on the YOLO family of one‐stage detectors. By constructing a carefully annotated dataset from the UTA-RLDD video corpus extracting image samples at 30 FPS, and verifying data diversity across gender, eyewear usage, illumination, and skin tone, we established a robust benchmark for evaluating multiple YOLO variants. Our systematic experiments demonstrated that the compact YOLOv11n model achieves a nearly state-of-the-art detection performance with minimal latency, making it especially well suited for embedded, on-board deployment. At the high-end of the spectrum, YOLOv9c attained the strongest overall accuracy, although with greater computational demands, positioning it as an ideal candidate when hardware resources permit cloud or accelerator support. We also implemented and assessed a classical Eye Aspect Ratio (EAR) approach, which offers a lightweight, CPU-only alternative capable of detecting sustained eye closure in real time. Our work provides a thorough comparative analysis of several YOLO architectures and highlights clear trade-offs between accuracy, inference speed, and hardware requirements. In future work, we plan to extend this framework by integrating temporal sequence modeling to capture the dynamics of drowsiness progression.

\bibliographystyle{splncs04}
\bibliography{mybibliography}

\begin{thebibliography}{10}
\providecommand{\url}[1]{\texttt{#1}}
\providecommand{\urlprefix}{URL }
\providecommand{\doi}[1]{https://doi.org/#1}

\bibitem{s22052069}
Albadawi, Y., Takruri, M., Awad, M.: A review of recent developments in driver drowsiness detection systems. Sensors,  \textbf{22}(5) (2022). \doi{10.3390/s22052069}  \textbf{22}(5) (2022). \doi{10.3390/s22052069}, \url{https://www.mdpi.com/1424-8220/22/5/2069}

\bibitem{nature2}
Arefnezhad, S., Hamet, J., Eichberger, A., Frühwirth, M., Ischebeck, A., Koglbauer, I.V., Moser, M., Yousefi, A.: Driver drowsiness estimation using eeg signals with a dynamical encoder–decoder modeling framework. Nature,  \textbf{12} (2022). \doi{10.1038/s41598-022-05810-x}  \textbf{12} (2022). \doi{10.1038/s41598-022-05810-x}, \url{https://rdcu.be/etXQH}

\bibitem{9929437}
Baccour, M.H., Driewer, F., Schäck, T., Kasneci, E.: Comparative analysis of vehicle-based and driver-based features for driver drowsiness monitoring by support vector machines. IEEE Transactions on Intelligent Transportation Systems,  \textbf{23}(12),  23164--23178 (2022). \doi{10.1109/TITS.2022.3207965}  \textbf{23}(12),  23164--23178 (2022). \doi{10.1109/TITS.2022.3207965}

\bibitem{8718312}
Budak, U., Bajaj, V., Akbulut, Y., Atila, O., Sengur, A.: An effective hybrid model for eeg-based drowsiness detection. IEEE Sensors Journal,  \textbf{19}(17),  7624--7631 (2019). \doi{10.1109/JSEN.2019.2917850}  \textbf{19}(17),  7624--7631 (2019). \doi{10.1109/JSEN.2019.2917850}

\bibitem{8520803}
Fujiwara, K., Abe, E., Kamata, K., Nakayama, C., Suzuki, Y., Yamakawa, T., Hiraoka, T., Kano, M., Sumi, Y., Masuda, F., Matsuo, M., Kadotani, H.: Heart rate variability-based driver drowsiness detection and its validation with eeg. IEEE Transactions on Biomedical Engineering,  \textbf{66}(6),  1769--1778 (2019). \doi{10.1109/TBME.2018.2879346}  \textbf{66}(6),  1769--1778 (2019). \doi{10.1109/TBME.2018.2879346}

\bibitem{9087013}
Garg, H.: Drowsiness detection of a driver using conventional computer vision application. In: 2020 International Conference on Power Electronics \& IoT Applications in Renewable Energy and its Control (PARC). pp. 50--53 (2020). \doi{10.1109/PARC49193.2020.236556}

\bibitem{ghoddoosian2019realisticdatasetbaselinetemporal}
Ghoddoosian, R., Galib, M., Athitsos, V.: A realistic dataset and baseline temporal model for early drowsiness detection (2019), \url{https://arxiv.org/abs/1904.07312}

\bibitem{9435480}
Hasan, F., Kashevnik, A.: State-of-the-art analysis of modern drowsiness detection algorithms based on computer vision. In: 2021 29th Conference of Open Innovations Association (FRUCT). pp. 141--149 (2021). \doi{10.23919/FRUCT52173.2021.9435480}

\bibitem{nature1}
Hassan, O., Ibrahim, A., Gomaa, A.: Real-time driver drowsiness detection using transformer architectures: a novel deep learning approach. Nature,  \textbf{15} (2025). \doi{10.1038/s41598-025-02111-x}  \textbf{15} (2025). \doi{10.1038/s41598-025-02111-x}, \url{https://rdcu.be/etXPO}

\bibitem{9509912}
Hussein, M.K., Salman, T.M., Miry, A.H., Subhi, M.A.: Driver drowsiness detection techniques: A survey. In: 2021 1st Babylon International Conference on Information Technology and Science (BICITS). pp. 45--51 (2021). \doi{10.1109/BICITS51482.2021.9509912}

\bibitem{10603125}
Qiu, S., Liu, D., Qin, Y., Tao, X.: Driver drowsiness detection using eeg and eog with an attention-cnn framework. In: 2024 4th International Conference on Computer Communication and Artificial Intelligence (CCAI). pp. 75--80 (2024). \doi{10.1109/CCAI61966.2024.10603125}

\bibitem{safarov2023real}
Safarov, F., Akhmedov, F., Abdusalomov, A.B., Nasimov, R., Cho, Y.I.: Real-time deep learning-based drowsiness detection: leveraging computer-vision and eye-blink analyses for enhanced road safety. Sensors,  \textbf{23}(14), ~6459 (2023)  \textbf{23}(14), ~6459 (2023)

\end{thebibliography}
%





\end{document}